\documentclass{article}

\PassOptionsToPackage{square,numbers}{natbib}
\usepackage[final]{neurips_2022}

\usepackage[utf8]{inputenc} %
\usepackage[T1]{fontenc}    %
\usepackage{hyperref}       %
\usepackage{url}            %
\usepackage{booktabs}       %
\usepackage{amsfonts}       %
\usepackage{nicefrac}       %
\usepackage{microtype}      %
\usepackage{xcolor}         %

\usepackage[pdftex]{graphicx} %
\usepackage{amsmath} %
\usepackage[super]{nth} %
\usepackage{numprint} %

\usepackage{multirow} 
\usepackage{siunitx}
\usepackage{etoolbox}           %
\renewcommand{\bfseries}{\fontseries{b}\selectfont} %
\robustify\bfseries             %
\newrobustcmd{\B}{\bfseries}    %
\newcommand{\itseries}{\fontshape{it}\selectfont} %
\robustify\itseries
\newrobustcmd{\IT}{\itseries} 

\title{Solving the Weather4cast Challenge \\
        via Visual Transformers for 3D Images}

\author{%
  Yury Belousov\thanks{Equal contribution} \\
  University of Geneva \\
  Switzerland \\
  \texttt{Yury.Belousov@unige.ch} \\
  \And
  Sergey Polezhaev\footnotemark[1]\\
  Neiro AI \\
  USA \\
  \texttt{sergey.polezhaev@outlook.com} \\
  \And
  Brian Pulfer\footnotemark[1] \\
  University of Geneva \\
  Switzerland \\
  \texttt{Brian.Pulfer@unige.ch} \\
}

\usepackage{xspace}
\newcommand{\ri}{$\operatorname{repeat\_interleave}$\xspace}
\newcommand{\SWINTR}{the SWIN-UNETR transformer\xspace}

\begin{document}

\maketitle

\begin{abstract}
    Accurately forecasting the weather is an important task, as many real-world processes and decisions depend on future meteorological conditions. The NeurIPS 2022 challenge entitled Weather4cast poses the problem of predicting rainfall events for the next eight hours given the preceding hour of satellite observations as a context. Motivated by the recent success of transformer-based architectures in computer vision, we implement and propose two methodologies based on this architecture to tackle this challenge. We find that ensembling different transformers with some baseline models achieves the best performance we could measure on the unseen test data. Our approach has been ranked \nth{3} in the competition.
\end{abstract}

\setcounter{footnote}{0} %
\section{Introduction}
The ability to predict the weather is of great importance in a variety of fields and applications where decision-making processes need to account for and heavily rely on upcoming meteorological conditions. Some examples include the transportation, energy, and agriculture industries \cite{nowcasting, Espeholt2022}. However, it is challenging to accurately predict the weather in the future, especially when the weather condition changes rapidly and unexpectedly. In addition to the traditional time-series data (e.g., air temperature, wind speed, precipitation, etc.), it also requires the consideration of the geographic context. The weather forecast in one region may have a strong correlation with that of other nearby regions \cite{meteored}. For example, heavy snow in the northeast of China may be related to light rain in the southwest of China. Weather forecasting requires the consideration of not only temporal correlations but also spatial correlations.

While recently a great amount of work has been focusing on weather nowcasting \cite{nowcasting, nowcasting2}, many socio-economic needs must be aware of future weather conditions that span several hours, and not just a couple. The Weather4cast challenge \cite{weather4cast_common} fills this gap by providing competitors with a tough challenge: predict rainfall events for the future eight hours given the preceding hour as a context.

In particular, the aim of the 2022 edition of the Weather4cast competition is to predict future high-resolution rainfall events from lower-resolution satellite radiances. While radar data is more precise, accurate, and of higher resolution than satellite data, they are expensive to obtain and not available in many parts of the world. We thus want to learn how to predict these high-value rain rates from radiation measured by geostationary satellites. Competition participants should predict rainfall locations for the next 8 hours in 32 time slots from an input sequence of 4 time slots of the preceding hour. The input sequence consists of four 11-band spectral satellite images. These 11 channels show slightly noisy satellite radiances covering so-called visible (VIS), water vapor (WV), and infrared (IR) bands. Each satellite image covers a 15-minute period and its pixels correspond to a spatial area of about 12km x 12km. The prediction output is a sequence of 32 images representing rain rates from ground-radar reflectivities. Output images also have a temporal resolution of 15 minutes but have a higher spatial resolution, with each pixel corresponding to a spatial area of about 2km $\times$ 2km. So in addition to predicting the weather in the future, participants also have to deal with a super-resolution task due to the coarser spatial resolution of the satellite data.

In contrast to the previous competition \cite{weather4cast_competition}, there is only one target variable --- rainfall events. The rainfall events data provided by the \textit{Operational Program for Exchange of Weather Radar Information (OPERA)\footnote{\url{https://www.eumetnet.eu/activities/observations-programme/current-activities/opera/}}} and the satellite images are given by \textit{European Organisation for the Exploitation of Meteorological Satellites (EUMETSAT)\footnote{\url{https://www.eumetsat.int/}}}.

Motivated by the recent success of transformer-based architectures for vision \cite{transformer, vit}, we decide to investigate them for the given task. In particular, we conduct experiments with the SWIN-UNETR transformer \cite{hatamizadeh2022SWIN} and an adaptation of the VIVIT \cite{vivit} model. Our results allow us to achieve \nth{3} place in the challenge and suggest that this is certainly an interesting research direction to pursue in future work.

\section{Experiments}

We present our experiments in four sub-sections. We first give a general overview of the architecture-independent concepts that can be applied to any model for this challenge. We then propose improvements to the baseline model provided with the challenge. Finally, we introduce our own approaches which interpret the task 4D input as a 2D video with a time dimension or as a 3D medical image with time as a depth dimension.

\subsection{Model-independent configurations}
\paragraph{Loss}
The selection of a proper loss function could be crucial for training. We try to optimize using the target competition's metric $\operatorname{IoU}$, a combination of Dice \cite{milletari2016vdice} and Focal \cite{lin2017focal} losses as well as a standard binary cross-entropy loss.

\paragraph{Dataset}
A discrepancy between the training and validation samples may result in a sub-optimal model choice. For instance, for one of the regions (\textit{roxi\_0007}) in 2020, the mean and maximum rain rates in training and validation sets differ by almost a factor of two ($\num{2.53e-2}$ and $\num{6.78e-2}$ for training, $\num{4.79e-2}$ and $\num{11.3e-2}$ for validation). To tackle this mismatch it is possible to treat the full validation set as part of the training one, increasing its size only by $840$ samples or less than $1\%$ (from $\numprint{228928}$ to $\numprint{229768}$ samples). Another possible solution is to calculate the difference mask and multiply it by the probabilities, thereby calibrating the model and making it less certain in places where rainfall prevailed in the training split and vice-versa.

\paragraph{Threshold}
When deciding the output class for a prediction of the model we need to choose a threshold value for its output above which the region is classified as rainy and vice-versa. The default value of $0.5$ could be sub-optimal, especially in the case of such an unbalanced dataset. We experimented with different threshold values ranging from $0.2$ to $0.7$.

\paragraph{Optimizer}
Following the previous winning solution \cite{leinonen2021improvements}, we conduct experiments with the $\operatorname{AdaBelief}$ \cite{zhuang2020adabelief} other than just the  $\operatorname{AdamW}$  \cite{loshchilov2017decoupledadamw} optimizer.

\paragraph{Temporal shift}
In order to specifically emphasize that the model predictions are time sequential, it is possible to predict the time deltas starting from the second time step. Specifically $t'_0 = t_0, t'_i = t'_{i - 1} + t_i \text{ for } i \ge 1$, where $t_i$ is a raw model's delta prediction from time $i-1$ to $i$ and $t'_i$ is the final prediction.

\subsection{BASELINE improvements}\label{subsection:baseline_improvements}
We explore several ways of improving the official baseline. These include using an attention grid \cite{oktay2018gridattention}, changing the activation from $\operatorname{RELU}$ to $\operatorname{RRELU}$ \cite{xu2015empiricalRRELU}, normalization from batch \cite{ioffe2015batch} to instance \cite{ulyanov2016instance}, and replacing transpose convolution with upsample and regular convolution \cite{odena2016deconvolution}.

\subsection{VIVIT}
Interpreting the challenge as a video-to-video classification task, we consider methods that perform well on video classification and video-to-video synthesis. Among all, we decide to focus on VIVIT \cite{vivit} and evaluate its performance.

Our VIVIT model applies two transformers: first, it uses a transformer on the spatial dimensions of the input, concatenating batch size and temporal dimension on the batch dimension. Then, for each sample of the batch, it uses the 4 obtained class tokens (from the 4 images constituting the training sample) to model the temporal information (extracting the time dimension out of the batch dimension) through a second transformer, which gives rise to a new final classification token for each sequence of images (i.e. input).

The final classification token needs to be used to obtain the final prediction, which is of much higher dimensionality. To this end, we reshape the token to obtain a squared tensor with height and width dimensions, and then we use linear interpolation to upscale the classification token to the desired spatial output size. A final convolution is applied to obtain the number of desired prediction steps.

\subsection{SWIN-UNETR transformer}
In this challenge, following the formulation, data points can be interpreted as 3D-shaped tensors instead. One of the few other fields where 3D data is common is Medical image processing. For this reason, we consider applying state-of-the-art techniques from the medical image analysis task. One of the recent successful and promising works dedicated to MRI image analysis is SWIN-UNETR \cite{hatamizadeh2022SWIN}. In this work, authors combined the UNETR \cite{hatamizadeh2022UNETR} architecture which performed well on different medical image segmentation tasks, and SWIN \cite{liu2021SWIN} transformers. SWIN is a hierarchical vision transformer and UNETR is built upon the U-Net-like architecture combined with vision transformers.

3D-shaped tensors with 4 channels are given as input to the SWIN-encoder. Firstly, this data is processed to create non-overlapping patches and then windowed for computing the self-attention. Later, these feature representations are fed to a convolutional decoder at several resolutions via skip connections.
To adapt this architecture for the weather prediction task, several techniques are being explored: \textbf{data transformation}, \textbf{upsampling}, and \textbf{channel convolution}.

\paragraph{Data transformation}\label{paragraph:data_transform}
To use SWIN-UNETR architecture almost "as-is", the training data needs to be transformed somehow. The model is not capable to do channel up-sampling and it is assumed that the input and output spatial dimensions are the same. Also since a size 2 and stride 2 pooling operation is applied 5 times in the original architecture, all spatial dimensions of the input image must be divisible by 32.

To overcome these issues, the height and width of the images are interpolated to size 256. Similarly, 4 time channels are repeated using PyTorch's \cite{pytorch} \ri operation to get 32 channels. Intuitively, this approach fits well to the task as, in fact, the model needs upsampling only because of the inconsistencies of the input and target formats and isn't strongly required to learn weights of any upsample operation. This approach is used in all SWIN-UNETR experiments unless otherwise specified.

\paragraph{Upsample}
Another approach to overcome discrepancies between the original SWIN-UNETR task and the weather prediction task is to modify the network architecture. In this experiment, we feed the data using its original shape to the network (except the height and width dimensions that were interpolated to 256) and change the last 2 layers of the SWIN encoder so they have an identical number of input and output channels. Also, the convolutional decoder is altered so the last $\operatorname{ConvTranspose}$ layers are up-sampling sequentially to the desired output size of 32.

\paragraph{Channel convolution}
We also experiment with other ways of using the SWIN-UNETR architecture without modifying it. In particular, we substitute the \ri operation with a convolution block, while everything else remains unchanged with respect to the method described in \nameref{paragraph:data_transform}. The convolution module consists of 2 Convolutional layers: the first one has 4 input channels and 32 output channels with kernel size 3, and for the second one input and output channels are 32 with the same kernel size of 3. $\operatorname{RRELU}$ activation and Instance Normalization are applied after these 2 layers.

\paragraph{Gradient checkpointing, mixed precision}
Some experiments require a comparatively large amount of GPU memory to train, therefore to be able to fit more training samples on the GPU we use gradient checkpointing. Also, mixed precision is used for the same purpose. Both methods don't affect model convergence heavily.

\section{Results}
\begin{figure}
	\centering
	\includegraphics[width=1.\linewidth]{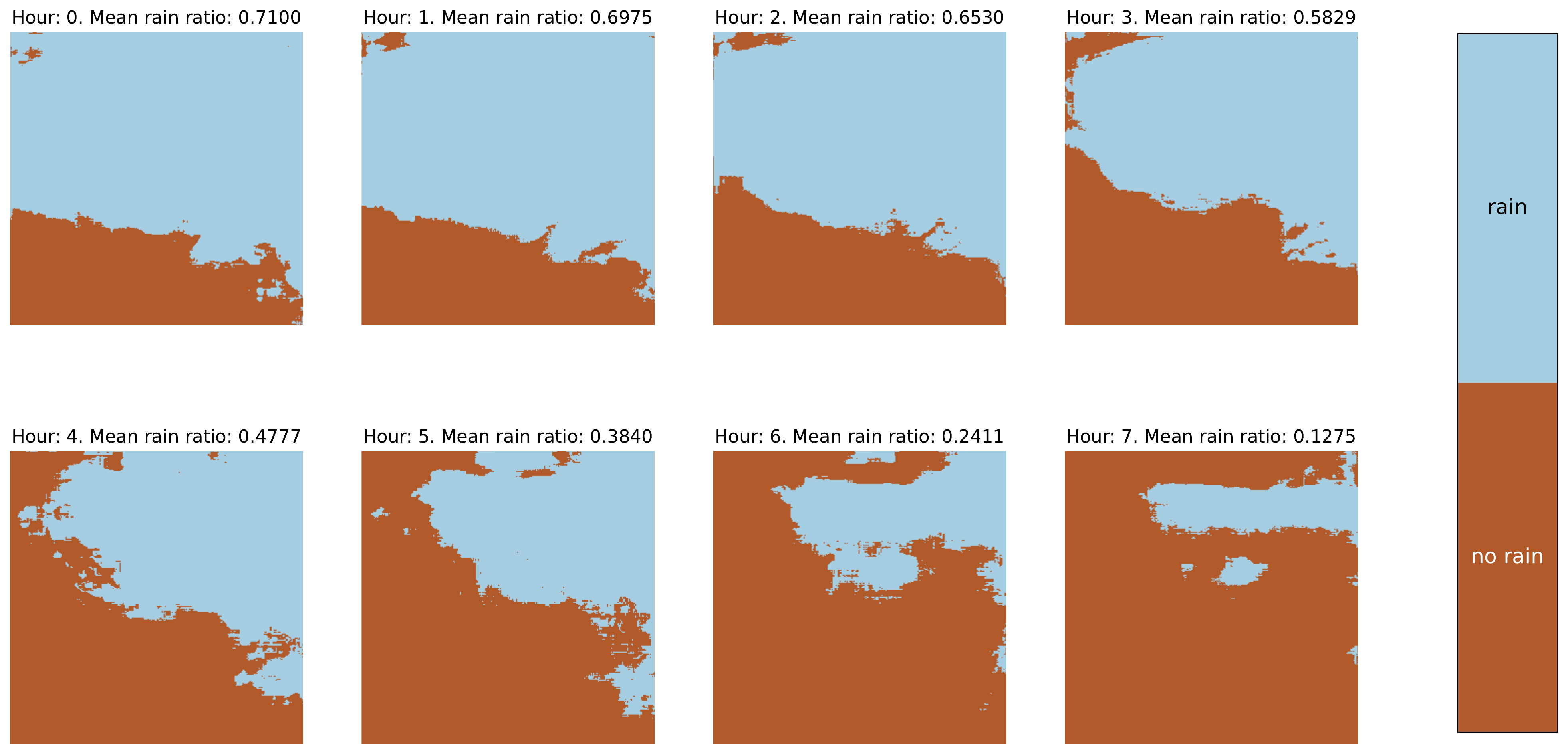}
	\caption{Example of model predictions for region \textit{roxi\_0004} in \textit{2020}. Each subplot's title contains a number of hours from the start of the prediction as well as an average rain ratio at the current time slot. The figure clearly shows the rain front gradually leaving this region as time passes.}
	\label{fig:model_prediction}
\end{figure}
{
\renewcommand{\arraystretch}{1.4}
\sisetup{detect-all=true,detect-mode,round-mode=places,detect-weight=true, round-precision=3}
\begin{table}
	\caption{Submissions to the test leaderboard. The official BASELINE is in \textit{italic}. \textbf{Bold} font is used to highlight the overall best mean performance per column.}
	\label{table:test}
	\centering
	\begin{tabular}{cccSSS}
		\toprule
		\textbf{Model type} & \textbf{Loss}      & \textbf{Submission name} & \textbf{Total mean} & \textbf{2019 mean} & \textbf{2020 mean} \\
		\midrule

		\multirow{7}{*}{BASELINE} & bce        & \IT official BASELINE                     & \IT 0.21267065  & \IT 0.241103729 & \IT 0.184237571 \\
                          & IoU        & Epoch 25                              & 0.190387623 & 0.209845814 & 0.170929431 \\
                          & bce        & Epoch 23                              & 0.213038092 & 0.243132971 & 0.182943213 \\
                          & bce        & train all. Epoch 24                   & 0.221893316 & 0.251573043 & 0.19221359  \\
                          & bce        & train all. Epoch 53                   & 0.166404013 & 0.185476943 & 0.147331083 \\
                          & bce        & improved, train all. Epoch 15          & 0.245426586 & \B 0.274123429 & 0.216729744 \\
                          & bce        & w/o convtranspose, train all. Epoch 33 & 0.246025382 & 0.267476343 & 0.224574421 \\
                          \midrule
\multirow{9}{*}{SWIN-UNETR}     & IoU        & Epoch 2                               & 0.18991962  & 0.206142471 & 0.173696769 \\
                          & dice focal & Epoch 2                    & 0.210168431 & 0.228120757 & 0.192216104 \\
                          & bce        & Epoch 3                               & \B 0.251699793 & 0.262365729 & 0.241033857 \\
                          & bce        & Epoch 3.  0.2 threshold               & 0.22744906  & 0.248051957 & 0.206846163 \\
                          & bce        & Epoch 3.  0.65 threshold               & 0.1937764  & 0.204287 & 0.183266 \\
                          & bce        & 16bit training. Epoch 3                        & 0.251606014 & 0.252863329 & \B 0.2503487   \\
                          & bce        & train all. Epoch 4                    & 0.249075743 & 0.261337657 & 0.236813829 \\
                          & bce        & channel conv. Epoch 1                 & 0.243861547 & 0.257976986 & 0.229746109 \\
                          & bce        & upsample. Epoch 1                    & 0.223967761 & 0.256281829 & 0.191653693 \\
                          \midrule
VIVIT                     & bce        & VIVIT base                    & 0.193286091 & 0.213283786 & 0.173288396        \\
		\midrule
		                          \multicolumn{3}{c}{take best prediction per region}                           & 0.290644         & 0.303460	        & 0.277827          \\
		\bottomrule
	\end{tabular}
\end{table}
}
\subsection{Evaluation}
In the core challenge, a model should predict the probability of rainfall events, trained on data across 7 regions and 2 years. The shape of an input to a model is $(11, 4, 252, 252)$, where $11$ is the number of bands spectral satellite images, $4$ is the time dimension ($1 \text{ preceding hour} \times 4 \text{ step}$, i.e. evenly divided into slots of $15$ minutes each) and $252 \times 252$ is the shape of a satellite region. Each prediction should have the following shape: $(32, 252, 252)$, where $32$ is the time dimension ($8 \text{ next hours} \times 4 \text{ step}$ with the same time discretization) and $252 \times 252$ is the shape of a rainfall region. Although the dimensions of the regions for input and output are the same, the spatial resolution of the satellite images is about six times lower than the resolution of the ground radar. This means that the  entire rainfall ground radar region resembles only a  $42 \times 42$ center's patch in the coarser satellite resolution, making this task a super-resolution task too. The remaining area should provide additional information to allow a model to predict rainfall not only for the current moment but also for the near future.
An example of a model's prediction for one of the regions is shown in \autoref{fig:model_prediction} (with  a time step of one hour, meaning that there are three more predictions between adjacent images that have not been displayed to save space). In total, there were $60$ different predictions for each region, totaling $60 \times 7 \text{ regions} \times 2 \text{ years} = 840$ predictions per submission in total.

\subsection{Metric}
The core metric used in this challenge is the intersection over union, which is defined as:
$\displaystyle\operatorname{IoU}(\mathcal{P},\mathcal{G})=\frac{|\mathcal{P}\cap\mathcal{G}|}{|\mathcal{P}\cup\mathcal{G}|}$
where \(\mathcal{P}\) and \(\mathcal{G}\) are the predicted and ground truth rainfall events, respectively.

For the second stage of the competition, the rain rate threshold was changed from 0 to 0.2, which is meteorologically more meaningful but increases the sparsity of the events to be predicted, making it a harder challenge and resulting in lower scores on the leaderboard.

\subsection{Test core partition}
The test section was available for one month. The teams were not limited by the total number of submissions, however, there was a limit of 5 concurrent submissions. The submissions' scores for this part are shown in the \autoref{table:test}. Due to the format of the competition and the time and resource limits, a full ablation study, as well as a complete examination of each of the improvements individually, remains open for future research. However, even from the current results, some conclusions can be made:
\begin{itemize}
    \item Counter-intuitively, optimizing the target metric ($\operatorname{IoU}$) directly leads to worse results, and this holds true for both baseline and SWIN-UNETR-based models. Using other loss functions, such as a combination of Dice \cite{milletari2016vdice} and Focal \cite{lin2017focal} losses, also does not improve performance. The best results were obtained using binary cross-entropy loss, with the weight of positive examples computed from the training set.
    \item Adding a validation dataset slightly improves the overall metric, but comes at the cost of making the validation metric inconsistent. Thus, it is no longer possible to rely on this score for model selection or early stopping, which may lead to overfitting.
    \item Baseline improvements from \autoref{subsection:baseline_improvements} tend to boost results. 
    \item VIVIT model is less competitive than the SWIN-UNETR counterpart. We suspect this has to do with how the final classification token, which gives a high-level representation of the observed state, is used to obtain the final prediction. 
    \item We do not observe much difference in terms of the final score when training with 16-bit precision.
    \item Both lowering or increasing the threshold for rain detection results in poorer performance. 
    \item Surprisingly, the simplest approach with \ri shows itself to be the best for \SWINTR.
    \item While modifying the network architecture for SWIN-UNETR performs well for some of the regions, it requires much more GPU memory for training and is considered too computationally expensive to conduct more experiments with it.
    \item Both \SWINTR and improved baseline perform significantly better than the official baseline.
    \item On average, the improved baseline performs better for the year 2019 while SWIN-UNETR is superior in 2020.
\end{itemize}

\subsection{Heldout core partition}
{
\renewcommand{\arraystretch}{1.4}
\sisetup{detect-all=true,detect-mode,round-mode=places,detect-weight=true, round-precision=3}
\begin{table}
	\caption{Submissions to the heldout leaderboard. The official BASELINE is in \textit{italic}. \textbf{Bold} font is used to highlight the overall best mean performance per column.}
	\label{table:heldout}
	\centering
	\begin{tabular}{cSSS}
		\toprule
		\textbf{Submission name}           & \textbf{Total mean}   & \textbf{2019 mean}     & \textbf{2020 mean}     \\
		\midrule

		\IT official BASELINE         & \IT 0.25508           & \IT 0.25902            & \IT 0.25115            \\

            BASELINE bce improved. Epoch 15 & 0.2696149372857143    & 0.26109335714285714    & 0.27813651742857143    \\
		SWIN-UNETR bce. Epoch 3           & 0.28139030064285714   & 0.2828801857142857     & 0.2799004155714286     \\

		majority vote                & \B 0.2997844244285715 & \B 0.29649640000000005 & \B 0.30307244885714285 \\

		\midrule

		take best prediction per region                 & 0.3024304015714286    & 0.30143774285714287    & 0.3034230602857143     \\
		\bottomrule
	\end{tabular}
\end{table}
}
The heldout section was available only for 4 days with a limit of 3 submissions per team in total, making the choice of model and weights a crucial issue. The submissions' scores for the heldout split are shown in the \autoref{table:heldout}. Once again, both \SWINTR and improved baseline perform particularly better than the official baseline. 

\subsection{Majority vote}
One way of solving the problem of choosing the optimal model is the majority voting approach. Namely, predictions of different models are generated, and the final prediction for each pixel is determined by the most frequent option. That is, if most models predict that it will rain at a given place at a given moment, that will be the final prediction and vice-versa.  We use the improved baseline and \SWINTR versions with \ri and $\operatorname{channel\_conv}$, trained with $\operatorname{AdamW}$ and $\operatorname{AdaBelief}$ optimizers. This method significantly outperforms each of the models individually.

\subsection{Best prediction per region}
As not only one overall metric but also values per region are shown to participants, a straightforward way to improve the score is to take the best predictions for each region individually and combine them into a new submission. This approach is shown in the last row of both \autoref{table:test} and \autoref{table:heldout}, and it reaches the \nth{3} place on the test leaderboard and  the ex-aequo \nth{3} place on the heldout leaderboard.

\section{Conclusion}

In this paper, we presented how we tackled the Weather4Cast competition. We first introduce a set of configurations that can be applied and enhance results for almost any model as well as baseline-specific improvements. We then propose two approaches based on Vision Transformers: the interpretation of the input as a video or as a 3D medical image and the use of problem-specific VIVIT or SWIN-UNETR Transformer respectively. As in previous years, we report that ensembling different trained models yield the most competitive results. Our selected algorithm has placed ex-aequo \nth{3} in the competition finals, which suggests that the use of transformers for weather prediction is a promising research direction that needs more investigation. 

The code for our solution is available at \url{https://github.com/bruce-willis/weather4cast-2022}.

\bibliography{common}

\end{document}